\documentclass{article}
\usepackage{ijcai23}

\usepackage{times}
\usepackage{soul}
\usepackage{url}
\usepackage[hidelinks]{hyperref}
\usepackage[utf8]{inputenc}
\usepackage[small]{caption}
\usepackage{graphicx}
\usepackage{amsmath}
\usepackage{amsthm}
\usepackage{booktabs}
\usepackage{algorithm}
\usepackage{algorithmic}
\usepackage[switch]{lineno}

\usepackage{url}
\usepackage{amsmath}
\usepackage{amssymb}
\usepackage{multirow}
\usepackage{algorithm}
\usepackage{algorithmic}
\usepackage{enumitem}
\usepackage{amsthm}
\usepackage{bm}

\newtheorem{theorem}{Theorem}

\newcommand{\Pt}{P_\text{t}}
\newcommand{\Ps}{P_\text{s}}

\def\argmin{\operatornamewithlimits{argmin}}

\newcommand{\myeqref}[1]{\ref{#1}}
\usepackage{booktabs}
\usepackage[stable]{footmisc}
\newtheorem{lemma}{Lemma}

\title{Learning Calibrated Uncertainties for Domain Shift: \\ A Distributionally Robust Learning Approach}

\author{
Haoxuan Wang$^1$
\and
Zhiding Yu$^2$\and
Yisong Yue$^3$\and
Animashree Anandkumar$^3$\and \\
Anqi Liu$^4$\footnote{Correspondence authors\protect\label{corres}.} \And
Junchi Yan$^1$\footref{corres}
\affiliations
$^1$Department of Computer Science and Engineering and MoE Key Lab of Artificial Intelligence, Shanghai Jiao Tong University \\
$^2$NVIDIA \\
$^3$Department of Computing and Mathematical Sciences, California Institute of Technology\\
$^4$Department of Computer Science, Johns Hopkins University
\emails
\{hatchet25, yanjunchi\}@sjtu.edu.cn,
zhidingy@nvidia.com,
aliu@cs.jhu.edu, \\ 
\{yyue, anima\}@caltech.edu
\footnote{Code repository: https://github.com/hatchetProject/Deep-Distributionally-Robust-Learning-for-Calibrated-Uncertainties-under-Domain-Shift
}
}

\begin{document}

\maketitle

\begin{abstract}
    We propose a framework for learning calibrated uncertainties under domain shifts, where the source (training) distribution differs from the target (test) distribution. We detect such domain shifts via a differentiable density ratio estimator and train it together with the task network, composing an adjusted softmax predictive form concerning domain shift. In particular, the density ratio estimation reflects the closeness of a target (test) sample to the source (training) distribution. We employ it to adjust the uncertainty of prediction in the task network. This idea of using the density ratio is based on the distributionally robust learning (DRL) framework, which accounts for the domain shift by adversarial risk minimization. We show that our proposed method generates calibrated uncertainties that benefit downstream tasks, such as unsupervised domain adaptation (UDA) and semi-supervised learning (SSL). On these tasks, methods like self-training and FixMatch use uncertainties to select confident pseudo-labels for re-training. Our experiments show that the introduction of DRL leads to significant improvements in cross-domain performance. We also show that the estimated density ratios align with human selection frequencies, suggesting a positive correlation with a proxy of human perceived uncertainties.
\end{abstract}

\section{Introduction}
Uncertainty estimation is an important machine learning problem that is central to trustworthy AI~\cite{10.1145/1085777.1085780,tomsett2020rapid}. In addition, many important downstream applications rely on the correct estimation of uncertainties. This includes unsupervised domain adaptation~\cite{zou2019confidence} and semi-supervised learning~\cite{sohn2020fixmatch}, where they are used to solicit confident pseudo-labels for re-training. In these applications, reliable pseudo-labels help avoid error propagation and catastrophic failures in early iterations~\cite{kumar2020understanding}.

Obtaining reliable uncertainty estimation is challenging. In contrast to human annotations of labels, obtaining the ground-truth uncertainties from real-world data can be costly or even infeasible. It is also known that the commonly used uncertainty proxies in deep neural networks, such as the softmax output, tend to give overconfident estimates~\cite{guo2017calibration}. This overconfidence is further amplified under domain shifts, where the target (test) domain and the source training domain differ significantly. Such distribution shifts tend to aggravate the existing issues in uncertainty estimation, leading to wrong but overconfident predictions on unfamiliar samples~\cite{li2020improving}.

Many methods have been proposed to calibrate the confidence of deep learning models so that the uncertainty level of a model prediction reflects the likelihood of the true event~\cite{guo2017calibration}. Label smoothing is a popular approach to reduce overconfidence and to promote more uniform outputs~\cite{szegedy2016rethinking}. Temperature scaling is another method where the logit scores are rescaled by a calibrated temperature~\cite{platt1999probabilistic}. Approaches such as Monte-Carlo sampling~\cite{gal2016dropout} and Bayesian inference~\cite{blundell2015weight} model uncertainties from a Bayesian perspective but are computationally expensive. Though these methods lead to more calibrated uncertainties, recent studies show that their results cannot be fully trusted under domain shifts~\cite{snoek2019can}.

\begin{figure*}[tb!]
\centering
\begin{tabular}{cc}
\centering
\includegraphics[height=4cm, trim={0cm, 0.1cm, 0cm, 0cm}, clip]{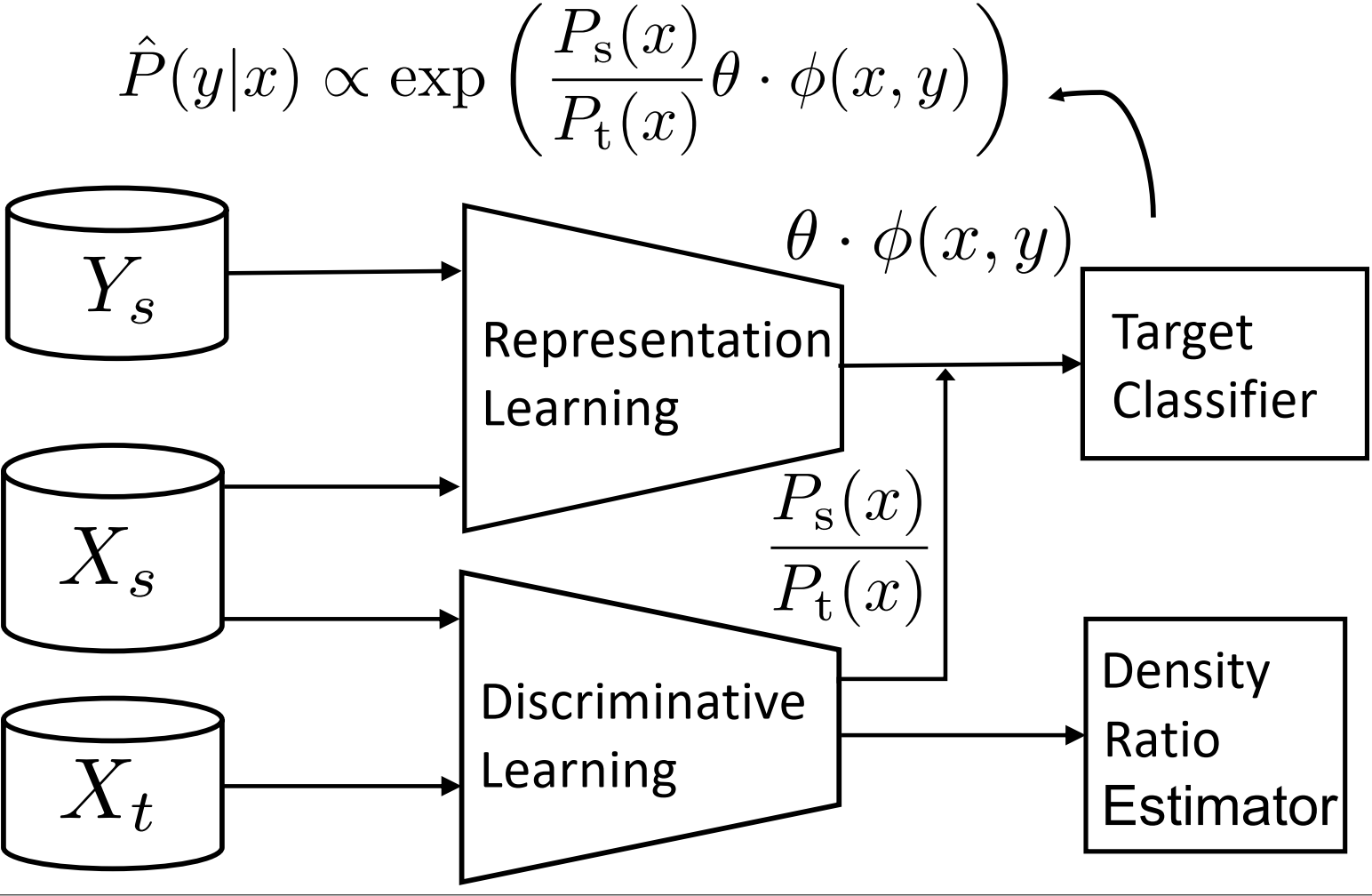}
&\includegraphics[height=4cm]{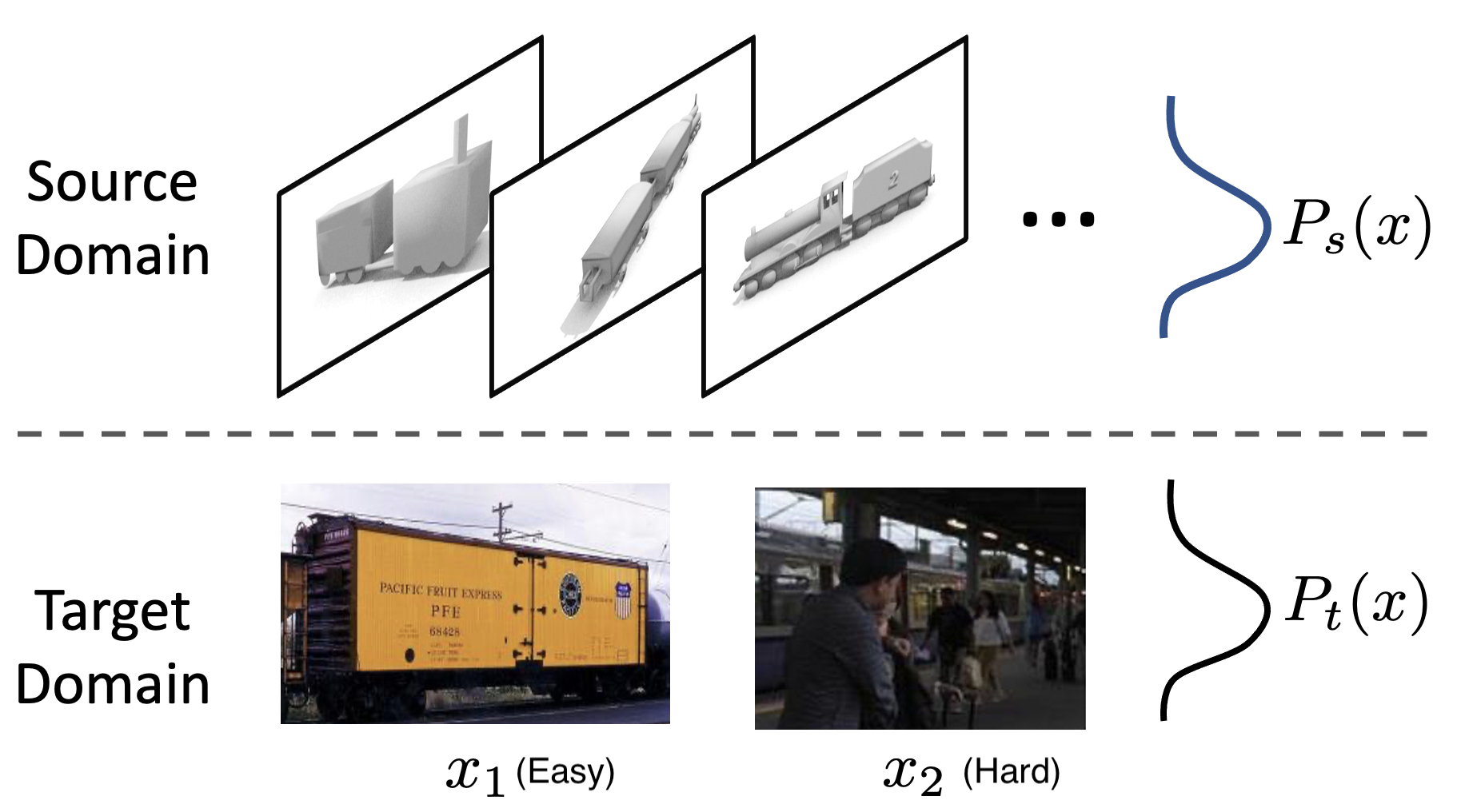} \\
(a) The end-to-end framework of DRL & (b) Density ratios: $\frac{P_s(\bm{x}_1)}{P_t(\bm{x}_1)}$ = 2.232, $\frac{P_s(\bm{x}_2)}{P_t(\bm{x}_2)}$ =  1.004 
\end{tabular}
\caption{(a) Architecture for end-to-end training of our DRL framework (see Sec.~\ref{sec:endtoend}). (b) Examples for category `Train' in VisDA. The estimated density ratios for the easy and hard target images are shown. The DRL framework gives higher uncertain predictions for the harder example ($\bm{x}_2$) that is more cluttered and hence not well-represented in the source domain.}
\label{fig:network}
\end{figure*}

 \paragraph{\bf Our approach:} To handle domain shifts, we characterize the ``overlap" between the source training data and the test data. Intuitively, if a test sample is distant from the training distribution, then its confidence level should be lowered. We incorporate this insight by estimating a density ratio for each sample and employ it for confidence calibration.

To be concrete, recall that the probability output for a classification neural network can be expressed as $P(y|\bm{x}) \propto \exp \left(\bm{\theta}_y \cdot \bm{\phi}(\bm{x})\right)$, where $\bm{\phi}(\bm{x})$ is the data feature for input $\bm{x}$, and $\bm{\theta}_y$ is the model parameter of the $y$-th class. We instead propose the following predictive form for our neural network:
\begin{equation} 
P(y|\bm{x}) \propto \exp \left(\frac{P_{\text{s}}(\bm{x})}{P_{\text{t}}(\bm{x})}\bm{\theta}_y \cdot \bm{\phi}(\bm{x})\right),
\label{eq:dratio} 
\end{equation}
where $P_{\text{s}}(\bm{x})$ and $P_{\text{t}}(\bm{x})$ are the densities of a data sample under the source and target distributions, respectively. When a target sample is close to the source domain (large $P_{\text{s}}(\bm{x})/P_{\text{t}}(\bm{x})$), the prediction is confident. However, when a target sample $\bm{x}$ is far away from the source distribution (small $P_{\text{s}}(\bm{x})/P_{\text{t}}(\bm{x})$), the confidence is lowered and the prediction is closer to a uniform distribution. This intuition is analogous to incorporating a sample-wise temperature to adjust the confidence according to the closeness of a test sample to the training distribution.

\begin{table}[tb!]
    \centering
    \begin{tabular}{ccccc}
    \toprule
     &  Paradigm & Data Distribution  & Prior\\
    \midrule 
    RBA & Two-stage & Simple (Gaussian)  & Yes \\
    Ours & End-to-end & Complex (Image)  & No \\
    \bottomrule
    \end{tabular}
    \caption{Comparison with the most related~\protect\cite{liu2014robust}.}
    \label{tab:drl_compare}
\end{table}

Eq.~\ref{eq:dratio} is based on the distributionally robust learning (DRL) framework. DRL is an adversarial risk minimization framework that involves a two-player minimax game between a predictor and an adversary \cite{grunwald2004game}. While many previous DRL methods~\cite{liu2020robust,nakka2020chance} operate in low-dimensional spaces using kernel density estimators for the density ratio estimation, we develop a DRL method to scale up to real-world computer vision tasks, which is able to produce calibrated uncertainties under domain shift. {\bf The highlights of the paper are}:

\textbf{1)} For the first time to our best knowledge, we propose a DRL method for uncertainty estimation under domain shift. We introduce a density ratio estimator which learns to predict the density ratios between source and target domains (Fig.~\ref{fig:network}(a)). The density ratio estimator and the target classifier are trained simultaneously in an end-to-end fashion. We also introduce additional regularization to improve calibration performance. A comparison of our framework with the most related DRL work is shown in Table \ref{tab:drl_compare}.

\textbf{2)} We show that the estimated density ratio reflects the distance of a test sample from both training and test distributions (Fig.~\ref{fig:network}(b)). Our experiments further empirically show that these estimates are also well correlated with human selection frequencies, based on the  ground-truth labels in ImageNetV2.

    
\textbf{3)} We empirically show that the top-1 class predictions of DRL are more calibrated than empirical risk minimization and temperature scaling on Office31, Office-Home, and VisDA. We measure the level of calibration using expected calibration error (ECE), Brier Score and reliability plot.
    
\textbf{4)} We integrate our method as a plug-in module in downstream applications such as unsupervised domain adaption and semi-supervised learning, leading to significant improvements. For example, incorporating self-training with DRL leads to state-of-the-art performance on VisDA-2017 and a $6\%$ improvement on hard examples. Incorporating Fixmatch with DRL improves the original Fixmatch by a relative 17\% increase in accuracy under the cross-domain setting.

\section{Related Work}
 \paragraph{Domain classification for domain shift:}
When dealing with classification problems under distribution shifts, one prevalent way is to discriminate different distributions. A popular approach is importance weighting using density ratio estimation~\cite{sugiyama2012density}, which reweighs source samples using pre-estimated density ratios to match the target distribution. However, importance weighting is known to be of high variance and only few works cover high-dimensional data~\cite{khan2019deep,park2020calibrated}. 
Other methods use a classification network to differentiate the learned source and target representations (adversarially) to locate a common subspace~\cite{ganin2016domain,adda,long2018conditional}. 
These works do not analyze the uncertainty estimation problem under domain shift. Yet our density ratio generation method is different from the above approaches in both motivation and implementation. We learn the density ratios using DRL's predictive form for generating calibrated predictions, and helps with choosing better pseudo-labeled data from the target domain in downstream tasks.

 \paragraph{Uncertainty calibration in deep models:} Uncertainty calibration aims for matching the model output probability with the true frequency of that event~\cite{verifiedUC}, and is achieving increasing attention in deep learning~\cite{nixon2019measuring}.
 Popular methods include Bayesian deep learning~\cite{gal2016dropout} 
and temperature scaling~\cite{guo2017calibration}.
However, these methods are either computationally expensive or not designed for the domain shift setting. Methods that focus on calibration for uncertainties under domain shift either do not fundamentally change the uncertainty generation process ~\cite{han2019unsupervised,lee2020model} 
or were built upon the traditional importance weighting setup~\cite{park2020calibrated,wang2020transferable}. In this paper, we directly generate more calibrated uncertainties based on the DRL framework under domain shift without directly optimizing the calibration error. 

 \paragraph{Distributionally robust learning under domain shift:} Several different learning algorithms can be derived from the DRL framework~\cite{liu2014robust,fathony2016adversarial,liu2017robust}. When applied to domain shift cases, an important limitation of these works is that they need density ratios to be estimated beforehand. This requires concrete prior distribution knowledge of the target domain, making them unsuitable for complicated tasks. Our work instead estimates the density ratio along with the learning process and is applicable on real-world tasks. While using a min-max framework, our work is orthogonal to adversarial training, which perturbs the covariate variable~\cite{hu2018does,najafi2019robustness} 
and focuses on robustness against adversarial perturbations. Our work derives a novel predictive form from class-regularized DRL by explicitly solving the min-max game and integrates differentiable density ratio estimation in the end-to-end training process. 

\section{Distributionally Robust Learning}
\label{sec:method}
In this section, we first review the preliminaries of DRL (Sec.~\ref{sec:drl}), followed by a proposed variant of DRL with class regularization (Sec.~\ref{sec:class-reg}). We then also propose an instantiation of DRL with a differentiable density ratio estimation network (Sec.~\ref{sec:endtoend}). Finally, we show the applications of our framework in UDA and SSL tasks (Sec.~\ref{sec:method}).

\subsection{Preliminaries}
\label{sec:drl}

 \paragraph{Notations and definitions:} Denote the input and labels by random variables $X$ and $Y$, respectively. Use $\bm{x}\in \mathbb{R}^d$ and $\mathcal{X}$ to represent the realization and sample space of $X$. Our goal is to find a predictor 
$\bm{f}(\bm{x}):\mathbb{R}^d\mapsto\mathbb{R}^C$, where $\bm{x}\in\mathcal{X}, \bm{f}\in\mathbb{R}^C\cap\Delta$,
that is close to the true underlying $P_t(Y|X)$. Here, $d$, $C$ and $\Delta$ denote the input dimension, class number, and probabilistic simplex, respectively. We consider the problem with labeled data sampled from a source distribution $\Ps(X, Y)$ and unlabeled data sampled from a target distribution $\Pt(X)$, and use ${P}_s(X, Y)$ to represent the empirical source distribution. In this work, we consider an important form of domain shift with the covariate shift assumption $P_\text{s}(X) \neq P_\text{t}(X)$, $P_{\text{s}}(Y|X) = P_{\text{t}}(Y|X)$.

 \paragraph{Motivation:} Traditional empirical risk minimization (ERM) frameworks tend to fail under covariate shift since ERM empirically learns a predictor $\hat{P}_{\text{s}}(Y|X)$ from the finite source data that usually fails to generalize to the target distribution. DRL is proposed to overcome this issue, which can be formulated as a two-player adversarial risk minimization game \cite{grunwald2004game} with the predictor player minimizing a loss, while the adversary player maximizing it. The adversary is allowed to perturb the labels, subject to certain feature-matching constraints to ensure data-compatibility. 
 
 \paragraph{Formulation:} For covariate shift, DRL~\cite{liu2014robust} deals with the mismatch between the expected loss and the training data, and is defined on \textbf{target} distribution:
\begin{align}
\hat{P}_{\text{t}}(Y|X)=\argmin_{\bm{f}} \;
\max_{\bm{g} \in \bm\Sigma} \; \mathbb{E}_{\bm{x}\sim \Pt(X)}\mathcal{L}\left(\bm{f}(\bm{x}), \bm{g}(\bm{x})\right),
\label{eq:game}
\end{align}
where $\bm{f}(\bm{x}), \bm{g}(\bm{x}) \in \mathbb{R}^C$ are the conditional label distributions given an input $\bm{x}$. $\bm{\Sigma}$ is a constraint for $\bm{g}$ to ensure the invariant conditional label distribution under covariate shift, which we will formulate in Eq.~\myeqref{eq:constraints}. Both $\bm{f}$ and $\bm{g}$ are not parameterized yet. 
$\mathbb{E}_{x\sim\Pt}\mathcal{L}(\cdot)$ is an expected log loss on the \textbf{target} input:
\begin{align}
\mathbb{E}_{\bm{x}\sim\Pt(X)}\mathcal{L}(\bm{f}(\bm{x}), \bm{g}(\bm{x})) 
\triangleq  \mathbb{E}_{\bm{x}\sim\Pt(X)}[-\bm{g}(\bm{x})\cdot\log \bm{f}(\bm{x})].
\end{align}
$\bm{f}$ is the predictor player minimizing the loss function while $\bm{g}$ is the adversary maximizing the loss function. After solving this game, $\bm{f}$ is our estimate of $\hat{P}_t(Y|X)$, which we will use for the classification task on the target domain. 

Eq.~\myeqref{eq:game} is defined on the \textbf{target} domain only. How could a predictor be properly trained while there are no target labels available? The answer is that the adversary $\bm{g}$ is implicitly constrained by the \textbf{source} features. We use the following constraints to make sure that $\bm{g}$ is close to $P_s(Y|X)$:
\begin{align}
\bm\Sigma = \{\bm{g}|
{\textstyle\sum}_{i} g_y\bm{\phi}(\bm{x}_i)=
{\textstyle\sum}_{i}\mathbb{I}[y_i=y]\bm{\phi}(\bm{x}_i), \forall y\},
\label{eq:constraints}
\end{align}
where $\bm{x}_i \sim P_{s}(X)$, $g_y$ is the $y$-th dimension of $\bm{g}$ and $\bm{\phi}(\bm{x}_i)$ is the feature for $\bm{x}_i$. Eq.~\myeqref{eq:constraints} is a necessary but not sufficient condition for $\bm{g} = P_s(Y|X)$, thus serving as an implicit constraint for $\bm{g}$ to be close to the true $P_t(Y|X)$ under the covariate shift assumption. Given a predefined feature function $\bm{\phi}$, when the adversary perturbs the conditional label distribution, certain aggregate function of $\bm{\phi}$ on $\bm{g}$ should equal to the counterpart on the empirical source data.

 \paragraph{From DRL to density ratio (derivation of $\textit{\textbf{f}}$):}
When using the expected target log loss in Eq.~\myeqref{eq:game}, Eq.~\myeqref{eq:dratio} is derived by solving the predictor $\bm{f}$. Here we refer the derivation details to \cite{liu2014robust} but emphasize an important property of the prediction: representation-level conservativeness, which means the predictions are more certain for inputs closer to the source domain (larger $\Ps(\bm{x})/\Pt(\bm{x})$) and more uncertain when $\Ps(\bm{x})/\Pt(\bm{x})$ is small. This property reflects the model's ability to convey information about what it does not know through the model uncertainty.

\subsection{Class-regularized Distributionally Robust Learning}
\label{sec:class-reg}

Inspired by label smoothing~\cite{szegedy2016rethinking} and regularization of neural network outputs \cite{pereyra2017regularizing}, we further add class-regularization to the prediction form in Eq.~\myeqref{eq:dratio}. Instead of doing it in a post-hoc way, we incorporate class-regularization into the original DRL formulation. We propose to use a weighted log loss to penalize the high confidence in the adversary's label prediction:
\begin{align}
\hat{P}_{\text{t}}(Y|X)=&\argmin_{\bm{f}} \;
\max_{\bm{g} \in \bm\Sigma} \; \mathbb{E}_{\bm{x}\sim\Pt(X)}[-\bm{g}(\bm{x})\cdot\log \bm{f}(\bm{x})] \notag \\
&- r  \mathbb{E}_{\bm{x}\sim \Pt(X)} \left[\bm{y}\odot \bm{g}(\bm{x})\cdot \log \bm{f}(\bm{x})\right],
\label{eq:regularize_game}
\end{align}
where $\bm{y}$ is the one-hot class vector, $\odot$ is the element-wise product, and $r \in [0,1]$ is a hyper-parameter that controls the level of regularization. $\bm\Sigma$ here is the same as in Eq.~\myeqref{eq:game}. We call this formulation \textbf{class-regularized} distributionally robust learning.
Eq.~\ref{eq:regularize_game} is a convex-concave function in terms of $\bm{f}$ and $\bm{g}$. According to the strong duality, we switch the order of the $\min$ and $\max$. With a fixed $\bm{g}$, $\bm{f} = \bm{g}$ is the optimal solution of the inner $\min$ problem. So we have the following lemma
(refer to appendix for proof) :

\begin{lemma}
\label{lemma:maxent}
Eq.~\myeqref{eq:regularize_game} can be reduced to a regularized maximum entropy problem with the estimator constrained: 
\begin{align}
\max_{\bm{f} \in \Sigma} \;&
\mathbb{E}_{\bm{x}\sim\Pt(X)}[-\bm{f}(\bm{x})\cdot\log \bm{f}(\bm{x})] \notag \\
&- r\mathbb{E}_{\bm{x}\sim \Pt(X)}\left[\bm{y}\odot \bm{f}(\bm{x})\cdot\log \bm{f}(x)\right],
\label{eq:maxent}
\end{align}
where $\bm\Sigma$ is the same as in Eq.~\myeqref{eq:constraints}, meaning that $\bm{f}$ should be close to the empirical source $P_s(Y|X)$.
\end{lemma}

\begin{theorem}
\label{thm:para}
The solution of Eq.~\myeqref{eq:maxent} takes the form:
$\bm{f}_{\theta, r}(y|\bm{x}) \propto \exp \left(\frac{\frac{P_{\text{s}}(\bm{x})}{P_{\text{t}}(\bm{x})}\bm{\theta}_y \cdot \bm{\phi}(\bm{x}) + r \mathbb{I}(y)}{r \mathbb{I}(y) + 1}\right),$
where $\bm{\theta}$ represents the model parameters and $\mathbb{I}(y)$ is the $y$th dimension of the one-hot encoding $\bm{y}$.
\label{thm:para}
\end{theorem}
The proof of Theorem \ref{thm:para} (refer to appendix)
follows the same principles of deriving Eq.~\myeqref{eq:dratio}. Different from \cite{liu2014robust}, we use Theorem~\ref{thm:para} as the new prediction form. In training, $\bm{y}$ is the one-hot encoding of each class. In inference, we set $\bm{y}$ to be an all-one vector. 

\paragraph{Class-level regularization:} Hyperparameter $r$ adjusts the smoothness of $\bm{g}$'s label prediction in Eq.~\myeqref{eq:regularize_game}. It translates to the $r\bm{y}$ term in the prediction form. Intuitively, it increases the correct label's prediction logits when it is smaller than $1$ and decreases the logits when it is larger than $1$. Thus $r$ provides additional regularization and smoothness to the conservative prediction.

\subsection{Differentiable Density Ratio Estimation}
\label{sec:endtoend}

Estimating $P_{\text{s}}(\bm{x})/P_{\text{t}}(\bm{x})$ is challenging~\cite{sugiyama2012density}, especially in high-dimensional spaces. Usually an estimator that calculates the densities in advance is used. But these estimates are usually sub-optimal in practice due to the different downstream tasks' objectives. We propose an end-to-end framework where \textbf{the density ratio estimator is trained together with the target classifier}. We introduce the novel differentiable density ratio estimation before proposing a joint training loss and the parameter learning process.

 \paragraph{Differentiable density ratio estimation:} Based on the Bayes' rule, $P_{\text{s}}(\bm{x})/P_{\text{t}}(\bm{x})$ can be computed from a conditional domain classifier \cite{bickel2007discriminative}:
$\frac{P_{\text{t}}(\bm{x})}{P_{\text{s}}(\bm{x})} = \frac{P(\bm{x}|\text{t})}{P(\bm{x}|\text{s})}  =  \frac{P(\text{t}|\bm{x})P(\text{s})}{P(\text{s}|\bm{x})P(\text{t})}$.
Concretely, they can be estimated via binary classification using unlabeled source and target data  with $\frac{P(\text{s})}{P(\text{t})}$ as a constant relating to the number of source and target samples.  
On the other hand, we observe that $P_{\text{s}}(\bm{x})/P_{\text{t}}(\bm{x})$
can be a trainable weight for each sample, and can be updated via the training objective of DRL. Therefore, we propose to train a binary classifier as the density ratio estimator, using both the binary cross-entropy loss and DRL's objective loss to update the network's parameters. 

Our approach is different from existing methods (e.g. \cite{dvn}) in three aspects: First, our binary classification network is proposed for the goal of achieving calibrated uncertainties under domain shift; Second, the weights trained this way lose their original properties as density ratios, but still reflect the relation between the two domains; Third, our obtained density ratios are mathematically the inverse of other methods' density ratios due to the DRL formulation.

 \paragraph{Joint training loss:} Assume $\bm{\phi}(\bm{x}, \bm{w}_r)$ is the representation learning neural network with parameter $\bm{w}_r$. We further define $P_d(X)$ as the joint distribution of both source data and target data with their domain labels $D=\{\bm{d}(\bm{x})\}$. We denote $\bm{\tau}(\bm{x}, \bm{w}_d)=(\tau_s, \tau_t)$ (where $\tau_s + \tau_t = 1$) as the two dimensional probability output of the source and target domains from a domain classifier $\bm{\tau}$ with parameter $\bm{w}_d$. Our joint training loss is defined as:
\begin{align}
\min_{\bm{w_r}, \theta, \bm{w_d}} & \mathbb{E}_{\bm{x}\sim \Pt(X)}\left[-\bm{g}_t(\bm{x})\cdot\log \bm{f}(\bm{x}; \bm{w}_r, \bm{\theta}, \bm{w}_d) \right] \notag \\ &+ \mathbb{E}_{\bm{x}\sim P_d(X)}\left[-\bm{d}(\bm{x})\cdot\log\bm{\tau}(\bm{x}, \bm{w}_d) \right],
\label{eq:joint_loss}
\end{align}
where $\bm{g}_t(\bm{x})=P_t(Y|X=\bm{x})$ and $\bm{f}(\bm{x}; \bm{w}_r, \bm{\theta}, \bm{r}_d)$ takes the form in Theorem \ref{thm:para}:

\begin{align}
\bm{f}&(\bm{x}; \bm{w}_r, \bm{\theta}, \bm{r}_d)\propto \notag \\  &\exp\left(\left(\frac{\tau_s(\bm{x},\bm{w}_d)}{\tau_t(\bm{x},\bm{w}_d)}\bm{\theta} \cdot \bm{\phi}(\bm{x},\bm{w}_r) + r \mathbb{I}(y)\right)/(r \mathbb{I}(y) + 1)\right).
\label{eq:deep_form}
\end{align}

 \paragraph{Parameter learning:} Note that $\bm{g}_t$ in the first loss term of Eq.~\ref{eq:joint_loss} concerns the conditional label distribution on target which is assumed to be not available. However, with the DRL formulation, we can compute and evaluate the gradients of $\bm{\theta}$ and $\bm{\phi}(\bm{x}, \bm{w}_r)$ directly with a change of measure in the derivation of the gradients so that the gradients' calculation does not depend on the target distribution (see details in appendix).
We show this in Fig.~\ref{fig:network}(a) that the representation learning network $\bm{\phi}$ only uses source data as the input. In this way, the parameter learning process that \textbf{originally depend on source and target data distributions (loss function) are now only associated with the source data and labels (gradient)}. We then update $\bm{\theta}$ and $\bm{w}_r$ using the computed gradients and also directly back-propagate from the second loss term in Eq. \ref{eq:joint_loss} to update $\bm{w}_d$. Finally, we treat the densities as trainable variables and derive gradients for them from the first loss term (details shown in appendix).
By the Bayes rule, $\frac{P_s(\bm{x})}{P_t(\bm{x})} = \frac{\tau_s P(t)}{\tau_t P(s)}$, where $P(t)$ and $P(s)$ are the amount of unlabeled data from each domain during the training process. Since we use the same amount of source and target data in each batch, they are canceled out by following $\frac{P(s)}{P(t)} = 1$. Then $\bm{f}$ in Theorem \ref{thm:para} is reduced to Eq. \ref{eq:deep_form}. Therefore, besides the binary classification loss, the parameter $\bm{w}_d$ of the discriminative network is also trained with gradients from the first loss term. Algorithm \ref{alg:endtoend} shows the details.

\begin{figure}[tb!]
\centering
\includegraphics[height=3.1cm, trim={ 6.5cm, 3.5cm, 2cm, 9cm}, clip]{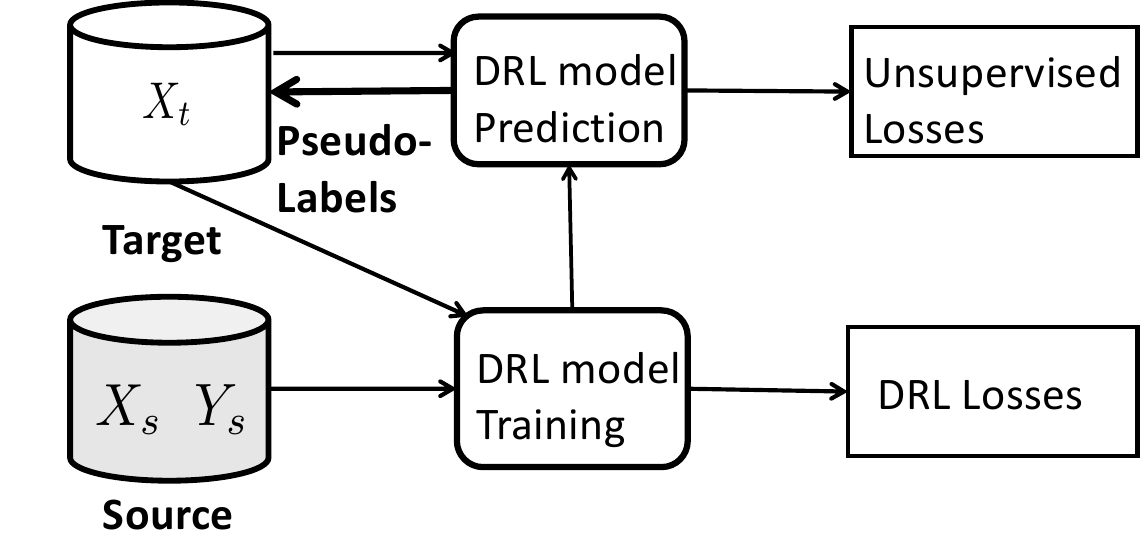}
\captionof{figure}{Formulation of the pseudo label based UDA or SSL methods with DRL. The unsupervised losses represent the loss imposed on the unlabeled target data. DRST conducts this procedure multiple iterations, while DRSSL minimizes the unsupervised losses on the augmented target data.}
\label{fig:abstract}

\end{figure}

 \paragraph{Applications to UDA and SSL:}
\label{sec:method}
We show how to incorporate end-to-end DRL within a framework that takes unlabeled data for training. We introduce the general setting with two examples: self-training based UDA and cross-domain SSL, followed by a new self-training algorithm DRST (distributionally robust self-training) and a new semi-supervised learning algorithm DRSSL (distributionally robust semi-supervised learning).

 \paragraph{General settings:} In many cases, the source domain may have abundant labels but the target domain lacks enough labels. Typical problems under this setting include unsupervised domain adaptation and semi-supervised learning. In both cases, a common strategy is to treat the prediction results on the target data as pseudo labels to train the model on target data. In these methods, model confidence (softmax output) is often leveraged as the proxy to rank the reliability of pseudo labels, with the underlying assumption that there exists a positive correlation between model confidence and pseudo label quality. However, such an assumption requires accurate uncertainty estimation to avoid false usage of wrong pseudo labels which may poison the training. To this end, we incorporate DRL into these frameworks in order to provide calibrated uncertainties. An illustration of this setting is shown in Fig.~\ref{fig:abstract}.

 \paragraph{Distributionally robust self-training:} In UDA, we are given labeled source data and unlabeled target data and aim to achieve adaptation from the source to the target domain. Self-training is an effective method for UDA, where the training procedure in Fig.~\ref{fig:abstract} is conducted multiple times. Here we propose DRST to plug the class-regularized DRL model into self-training. The idea is to regard each training epoch as a new domain shift problem in DRL. After each training epoch, we make predictions on the target domain and select more confident data and generate pseudo-labels for them to merge into the source training data. Both the pseudo-labels and the model confidence are achieved from Theorem \ref{thm:para}. Then the labeled source data and the newly pseudo-labeled target data become the new source set for the next training epoch. 

\begin{algorithm}[tb!]
\caption{End-to-end Training for DRL}
\label{alg:endtoend}
\begin{small}
\begin{algorithmic}[1]
\STATE {\bfseries Input}: DNN $\bm{\phi}$ and DNN $\bm{\tau}$, with optimizer SGD$_1$ and SGD$_2$, respectively. Learning rates $\gamma_1$ and $\gamma_2$, epoch number $T$.
\STATE {\bfseries Initialization}: $\bm{\phi}, \bm{\tau} \leftarrow$ random initialization, epoch $\leftarrow 0$
\STATE {\bfseries While} epoch $< T$
\STATE \qquad {\bfseries For} each data mini-batch
\STATE \qquad\qquad Update $\bm{\tau}$ by SGD$_1(\gamma_1)$ using the combined gradients from both loss terms in Eq. \ref{eq:joint_loss};
\STATE \qquad\qquad Compute $\bm{f}$ using $\bm{\theta}$, $\bm{w}_r$, and $\bm{w}_d$;
\STATE \qquad\qquad Update $\bm{\phi}, \bm{\theta}$ by SGD$_2(\gamma_2)$ using derived gradients;
\STATE \qquad epoch $\leftarrow$ epoch $+1$
\STATE {\bfseries Output}: Trained networks $\bm{\phi}$, $\bm{\tau}$.
\end{algorithmic}
\end{small}
\end{algorithm}

 \paragraph{Distributionally robust semi-supervised learning:} In cross-domain SSL where there is little labeled source data and much unlabeled target data, we aim to utilize unlabeled data in the target domain to help representation learning and save the effort needed for labeling. One effective strategy is to use pseudo-labels generated from weakly-augmented data to supervise strongly-augmented data. Here `weakly' means simple flip-and-shift data augmentation while `strongly' follows the same strategy as FixMatch~\cite{sohn2020fixmatch}. Using  DRL's prediction, we propose DRSSL, which assigns pseudo-labels more conservatively. Practically, we plug DRL into Fixmatch. Here the unsupervised loss in Fig.~\ref{fig:abstract} is the `consistency loss' in FixMatch:
  $  \mathcal{L}_u =  \frac{1}{M}\sum_{m=1}^{M} \mathbb{I}(\max(\hat{P}(y^w_t|x^w_t))>\eta) H(\hat{y}^w_t,\hat{P}(y^s_t|x^s_t) )$,
where $x_t^w$ and $y_t^w$ represent the weakly-augmented target data, $x_t^s$ and $y_t^s$ represent the strongly-augmented version of the same image data, and $\eta$ is a threshold for generating pseudo-labels $\hat{y}^w_t$.

\section{Experiments}
\label{sec:exp}
We evaluate our method on benchmark datasets and compare our performance with other uncertainty quantification, calibration, and domain adaptation baseline methods.
DRL is evaluated as a method providing more calibrated uncertainties (Sec. \ref{sec:drl_results}),  DRST as an effective UDA method (Sec. \ref{sec:drst_results}) and DRSSL as a cross-domain SSL method. We show additional results and details in the appendix. 

 \paragraph{Datasets and methods:} We use Office31~\cite{office31}, Office-Home~\cite{officehome} and VisDA2017~\cite{visda2017} for evaluating DRL's uncertainties. We compare DRL with temperature scaling (TS), VADA~\cite{shu2018dirt} and source-only. We also train models using ImageNet~\cite{2009ImageNet} as the source domain and ImageNetV2~\cite{2019imagenetv2} as the target domain to check the relationship between our estimated weights and the human selection frequencies (HSF)~\cite{chen2020angular}. 

\begin{figure*}[tb!]
\centering
\begin{tabular}{l}
\includegraphics[height=6.5cm, width=14cm, trim={0cm, 0.0cm, 0.0cm, 0.15cm},clip]{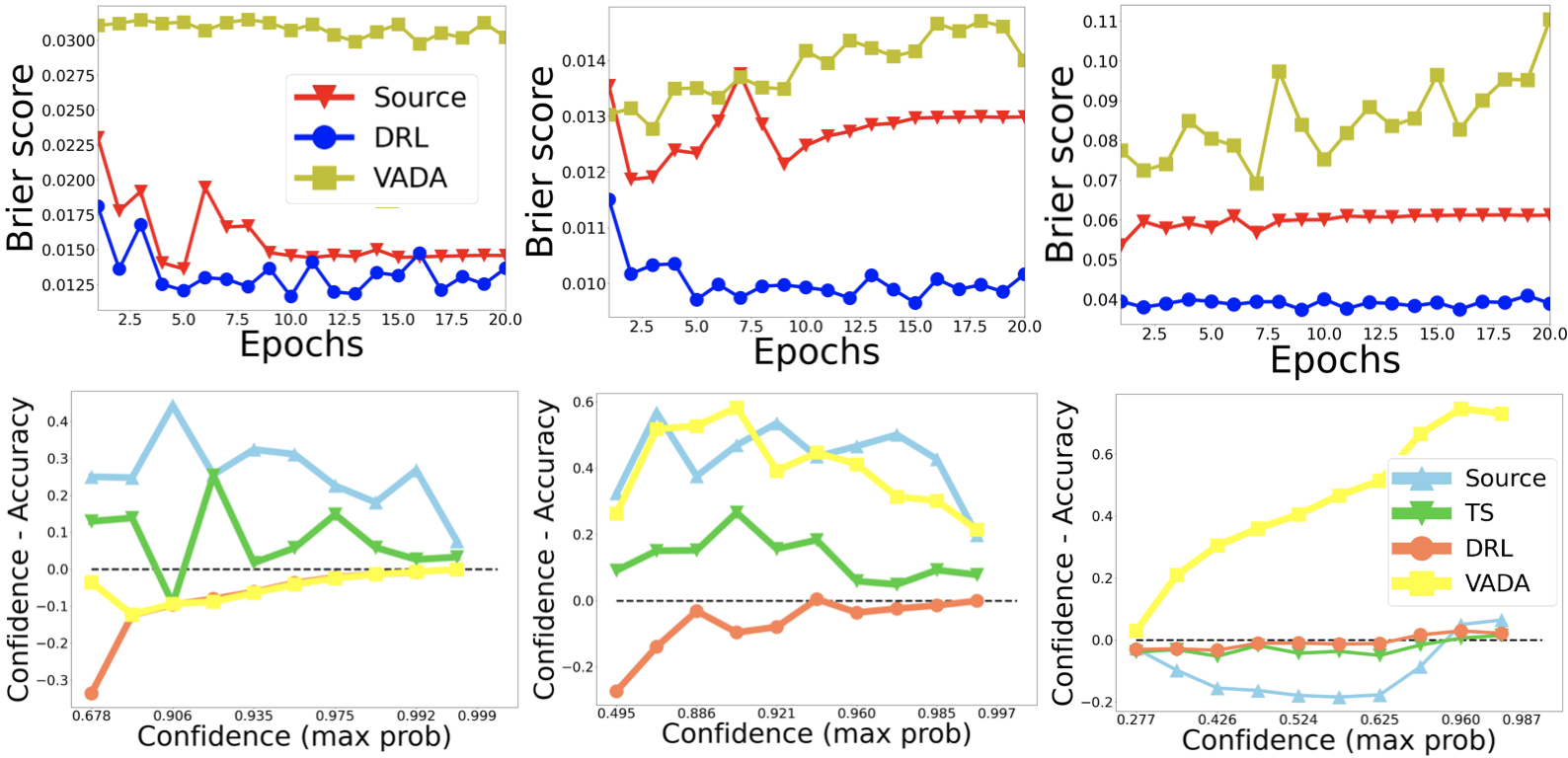} \\
 \quad \quad \quad  (a) Office31 (A$\rightarrow$W)  \quad \quad \quad \quad (b) Office-Home (P$\rightarrow$A)  \quad \quad \quad  \quad \quad \  (c) VisDA 
\end{tabular}
\caption{Brier score (top) and reliability diagrams (bottom) on \textbf{Office31}, \textbf{Office-Home} and \textbf{VisDA}. DRL generates more calibrated uncertainties than source-only and temperature scaling and VADA. Brier score measures the mean squared difference between the predicted probability and the actual outcome.  For a fully calibrated classifier, the confidence should match the accuracy across the full range of confidence. Thus the closer the lines are to the dashed line, the more calibrated the method is. Our method gets more and more calibrated as the confidence increases. Note that in the first row, TS's Brier scores are much larger and excluded to not effect the scale.}
\label{fig:boost}
\end{figure*}

VisDA2017 is also used to evaluate UDA performance. We compare with (1) traditional UDA baselines: 
MCD~\cite{saito2018maximum} and ADR~\cite{saito2018adversarial}; (2) self-training baselines: CBST~\cite{zou2018domain} and CRST~\cite{zou2019confidence}; (3) methods that tackle domain adaptation with uncertainty: BRER~\cite{han2019unsupervised} and MUDA~\cite{lee2020model}; (4) uncertainty quantification methods combined with self-training: AVH~\cite{chen2020angular}+CBST.
In addition, we use CIFAR10, STL10~\cite{stl10}, MNIST~\cite{mnist} and SVHN~\cite{svhn} to construct cross-domain SSL settings, which has few source labeled data and much unlabeled target data and show DRSSL's advantages in cross-domain SSL over Fixmatch~\cite{sohn2020fixmatch}.

    \begin{table}[tb!]
   \centering	  
\begin{tabular}{c|ccc}
\toprule
HSF & Source & Temp.Scal. & DRL (Ours) \\
\midrule
\text{[0.0, 0.2]} & 0.2694 & 0.2624 & \textbf{0.0129} \\ 
\text{[0.2, 0.4]} & 0.1818 & 0.1745 & \textbf{0.0036} \\ 
\text{[0.4, 0.6]} & 0.1344 & 0.1281 & \textbf{0.0012} \\ 
\text{[0.6, 0.8]} & 0.0667 & 0.0601 & \textbf{0.0019} \\ 
\text{[0.8, 1.0]} & 0.0319 & 0.0246 & \textbf{0.0019} \\
\bottomrule
\end{tabular}
\captionof{table}{Expected calibration error (ECE) comparison on ImageNetV2 under different HSF.}
\label{tab:imgnet_unc_score}
\end{table}

 \paragraph{Evaluation metrics:} Apart from accuracy, we also use Brier score~\cite{brier1950verification}, expected calibration error (ECE)~\cite{guo2017calibration}, and reliability plots~\cite{guo2017calibration} to evaluate the performance of our proposed method and the baselines. Brier score measures the mean squared difference between the predicted probability assigned to the possible outcome and the actual outcome. ECE is defined as the sum of average difference between prediction accuracy and confidence of different confidence bins (we use 15 bins in practice). Despite the potential problems of ECE~\cite{nixon2019measuring}, it is still the most prevalent metric for the top-1 prediction. For both the Brier score and ECE, the lower the score, the more calibrated is the model.

 \paragraph{Experimental setup:} For Office31 and Office-Home tasks, we use ResNet50~\cite{resnet} as the backbone for all models. We train with SGD for $100$ epochs and set the learning rate to $0.001$. For VisDA, we use ResNet101 and SGD optimizer. During the $20$ epochs of training, the initial learning rate is set as $10^{-5}$ and the weight decay parameter is set as $5\times 10^{-4}$. For ImageNet, we follow the standard training process of AlexNet~\cite{alexnet} and VGG-19~\cite{vgg19}, where the initial learning rate is $0.01$ and we decay the learning rate by a factor of 10 for every 30 epochs. All of the training are done on DGX V100 Tesla V100 GPUs with 32GB memory. The main packages and corresponding versions are: PyTorch 0.4.0, CUDA 10.1.

\begin{figure}[tb!]
\centering
\includegraphics[height=4cm, trim={1.0cm, 0.5cm, 2.4cm, 0cm},clip]{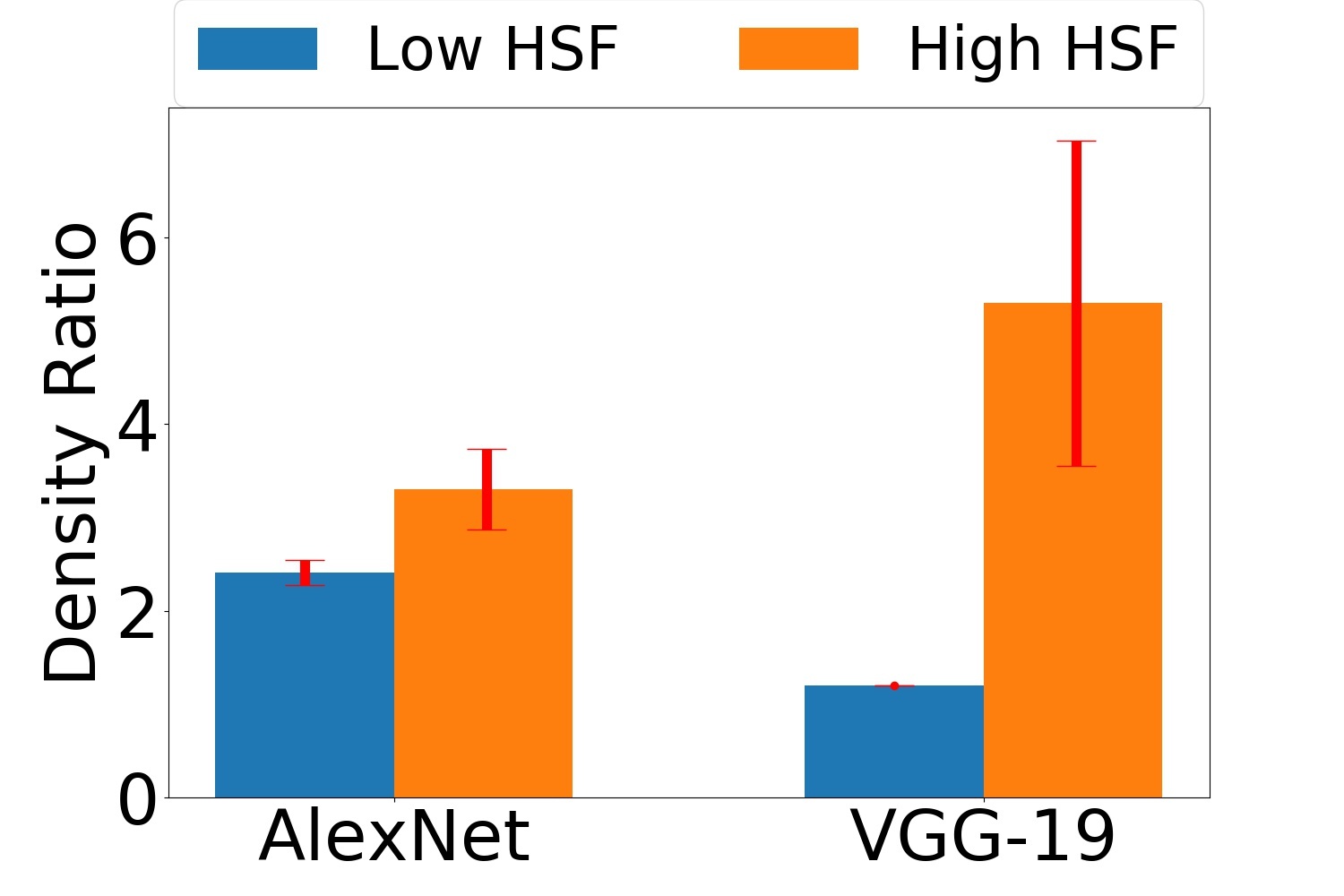}
\caption{Density ratios vs HSF on ImageNetV2.}
\label{fig:bar_plot}
\end{figure}

\subsection{Calibrated Uncertainties from DRL}
\label{sec:drl_results}


\begin{figure*}[tb!]
\centering
\begin{tabular}{ccc}
\includegraphics[height=3.8cm]{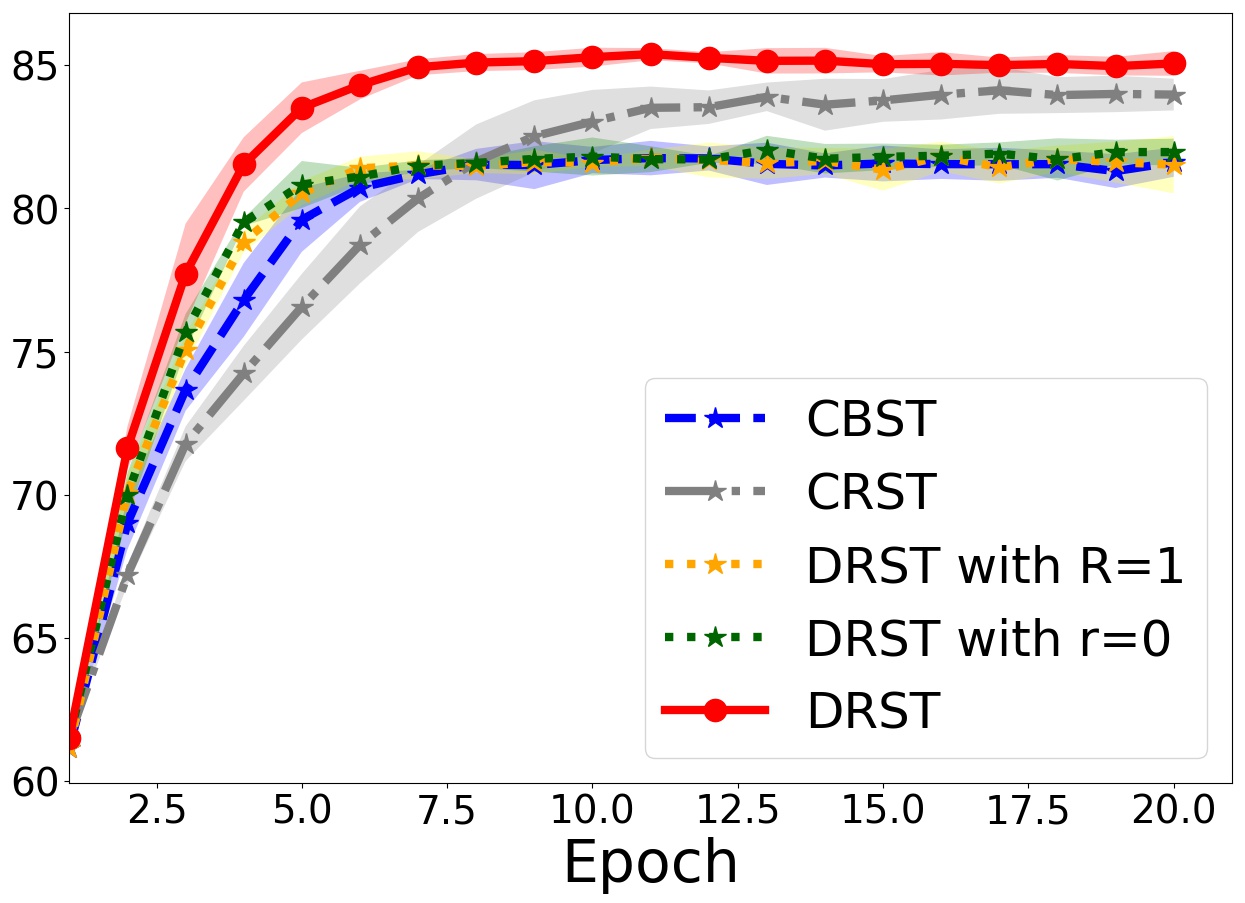}& 
\includegraphics[height=3.8cm]{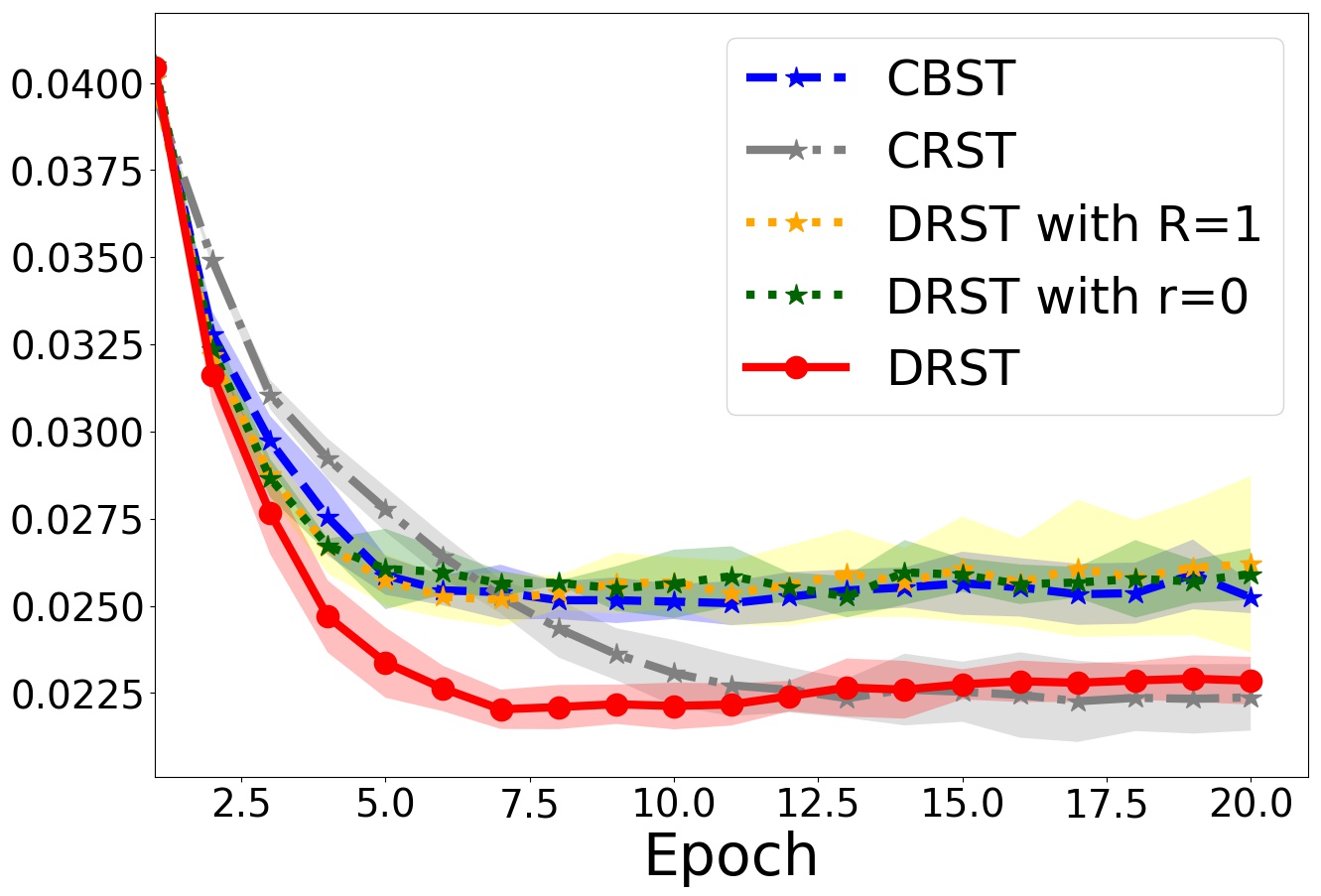} &
\includegraphics[height=3.8cm]{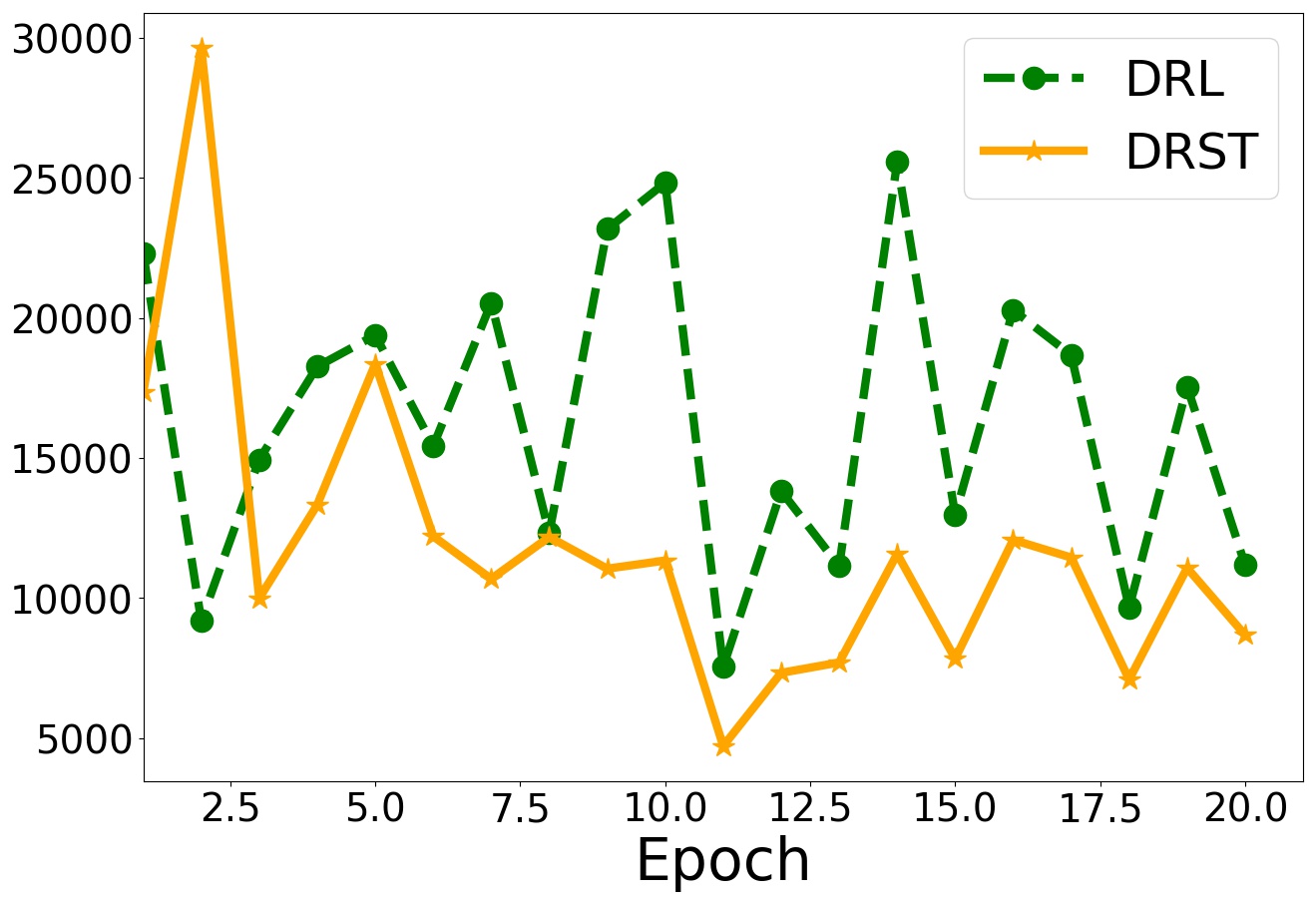} \\
(a) Test accuracy & (b) Brier score & (c) Covariate shift
\end{tabular}
\caption{(a)-(b) Results on VisDA-17 (performed with 5 random seeds) with test accuracy and Brier score. DRST outperforms the baselines significantly. (c) We adopt distribution gap $P_s(\bm{\phi}(\bm{x}))/P_t(\bm{\phi}(\bm{x})) - P_s(\bm{\phi}(\bm{x}), y)/P_t(\bm{\phi}(\bm{x}), y)$ as a proxy of covariate shift. DRST helps further reduce this gap with self-training. }
\label{fig:drst}
\end{figure*}

 \paragraph{DRL's calibrated confidence:} Fig.~\ref{fig:boost} shows our method achieves better uncertainty calibration. Note that the type of uncertainty we measure here is the epistemic uncertainty~\cite{gal2016dropout}, caused by the lack of data and domain gaps. The aleatoric uncertainty is caused by noise inherent in data and is out of our scope. DRL tends to be underconfident and conservative but still stays closer to the calibration line (dashed line). Table~\ref{tab:imgnet_unc_score} shows our method obtains better ECE on ImageNetV2 in different human selection frequency (HSF) bins. HSF is defined as the average number of times an image from a given class is classified correctly by a group of annotators. Say there are $n$ annotators and an image with ground-truth label ``car'',  $p$ annotators recognize the image as ``car'', then the image's HSF is $\frac{p}{n}$. The lower the HSF of an image, the harder for humans to correctly classify it. 
 Appendix
provides additional results and shows that the accuracy of DRL is also competitive.

 \paragraph{Density ratio v.s. HSF:} Fig.~\ref{fig:bar_plot} shows that images with low HSF (visually harder ones) have smaller density ratios, indicating that the estimated density ratios are positively correlated with HSF on ImageNetV2 under different network architectures. Here we regard the value of HSF from [$0, 0.2$] as low and $(0.2, 1]$ as high.

\subsection{Unsupervised Domain Adaptation}
\label{sec:drst_results}
 \paragraph{Initialization with ASG:} The pretrained models impact performance. Automated synthetic-to-real generalization (ASG) model~\cite{chen2020automated}, which uses ImageNet pretrained knowledge, can be used to improve synthetic training and self-training.
Hence, we use the ASG pretrained model as initialization, with results in Table \ref{tab:visda}'s last four rows.

 \paragraph{Accuracy and calibration:} Fig.~\ref{fig:drst}(a) and Table~\ref{tab:visda} show that DRST performs best in accuracy. Our vanilla version of DRST outperforms CRST by over 5\% with ASG initialization. We improve the SOTA self-training accuracy on VisDA by over 1\%. In the rest of the results, we use DRST to represent DRST-ASG. In comparison with uncertainty based methods (not shown in the table), BRER achieves an accuracy of 80.59±1.39\% while ours is 83.75\% using the same source model. MUDA aims to minimize model uncertainty and only yields a result of 78.5\%. We only identified two methods with higher accuracy on VisDA2017: CDTrans~\cite{Xu2021CDTransCT} (using transformer) and CPGA~\cite{Qiu2021SourcefreeDA} (two stage learning by calculating prototypes). Both of them cannot estimate uncertainty and it is nontrivial for them to incorporate uncertainty or pseudo-label modeling. Fig.~\ref{fig:drst}(b) shows that DRST also achieves better calibrated performance.

\begin{figure}[tb!]
\centering
\begin{tabular}{c}
\includegraphics[height=4cm]{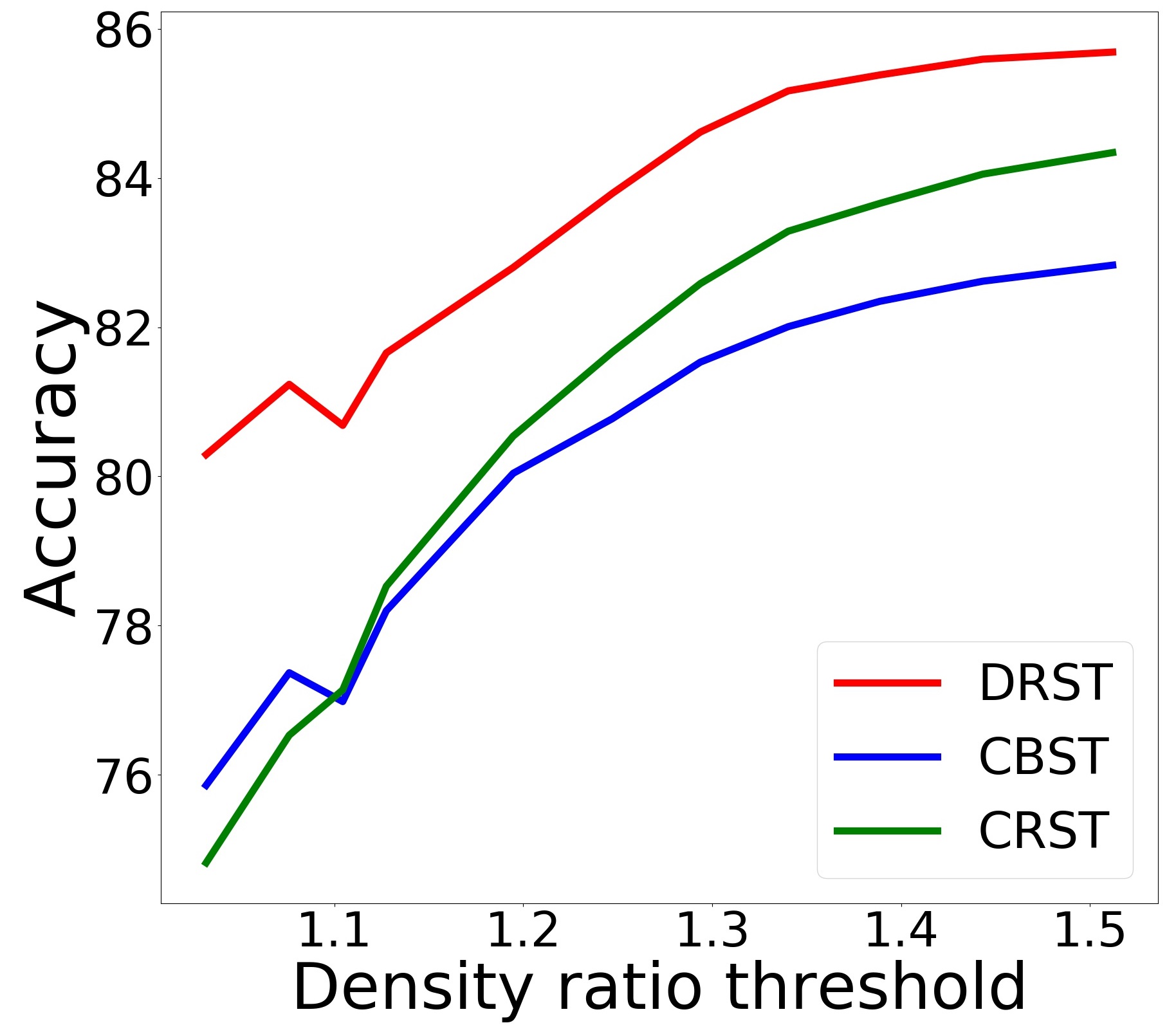}
\end{tabular}
\caption{Accuracy vs. estimated density ratio weights. Improvement from DRST increases on harder examples (lower weights).}
\label{fig:drst_d}
\end{figure}

\begin{table*}[tb!]
\centering
\resizebox{\textwidth}{23mm}{
\begin{tabular}{cccccccccccccc}
\toprule[2pt]
Method   & Aero  & Bike  & Bus   & Car   & Horse & Knife & Motor & Person & Plant & Skate & Train & Truck & Mean  \\
\midrule[2pt]
Source \cite{saito2018adversarial}  & 55.1  & 53.3  & 61.9  & 59.1  & 80.6  & 17.9  & 79.7  & 31.2   & 81.0  & 26.5       & 73.5  & 8.5   & 52.4  \\
MCD \cite{saito2018maximum}  & 87.0  & 60.9  & \textbf{83.7}  & 64.0  & 88.9  & 79.6  & 84.7  & 76.9   & 88.6  & 40.3       & 83.0  & 25.8  & 71.9  \\
ADR \cite{saito2018adversarial}     & 87.8  & 79.5  & \textbf{83.7}  & 65.3  & 92.3  & 61.8  & 88.9  & 73.2   & 87.8  & 60.0       & 85.5  & 32.3  & 74.8  \\
CBST \cite{zou2018domain} & 87.2  & 78.8  & 56.5  & 55.4  & 85.1  & 79.2  & 83.8  & 77.7   & 82.8  & 88.8       & 69.0  & \textbf{72.0}  & 76.4  \\
CRST \cite{zou2019confidence} & 88.0  & 79.2  & 61.0  & 60.0  & 87.5  & 81.4  & 86.3  & 78.8   & 85.6  & 86.6       & 73.9  & 68.8  & 78.1  \\
CBST-AVH \cite{chen2020angular}  & 93.3  & 80.2  & 78.9 & 60.9 & 88.4 & 89.7 & 88.9 & 79.6 & 89.5 & 86.8 & 81.5  & 60.0  & 81.5  \\ 
\textbf{DRST} (Ours) & 93.47 & 86.30 & 65.74 & 68.03 & 93.99 & 95.08 & 87.34 & 83.30 & 92.97 & 88.65 & 83.66 & 66.42 & 83.75 \\
\midrule
ASG \cite{chen2020automated}  & 88.81 & 68.55 & 65.31 & 78.06  & \textbf{95.78} & 9.11  & 84.89 & 29.58 & 82.13 & 33.76 & \textbf{86.00} & 12.04 & 61.17    \\
CBST-ASG \cite{chen2020automated} & \textbf{95.12} & \textbf{86.53} & 79.83 & 76.01 & 94.61 & 92.34 & 85.94 & 75.08  & 89.23 & 82.16      & 73.42 & 56.49 & 82.23 \\
CRST-ASG \cite{chen2020automated} & 92.38 & 81.30 & 74.63 & \textbf{84.40} & 90.90 & 92.43 & \textbf{91.65} & \textbf{83.78}  & \textbf{94.92} & 88.12      & 74.88 & 61.10 & 84.21 \\
\textbf{DRST-ASG} (Ours)     & 94.51 & 85.58 & 76.50 & 77.18 & 94.39 & \textbf{95.33} & 88.89 & 81.23  & 94.22 & \textbf{90.36}      & 81.75 & 63.10 & \textbf{85.25} \\
\bottomrule[2pt]
\end{tabular}
}
\caption{Accuracy comparison with different UDA and self-training methods on VisDA2017. ``Skate'' denotes ``Skateboard''.}
\label{tab:visda}
\end{table*}

 \paragraph{Ablation study:} Fig.~\ref{fig:drst}(a)(b) include two ablation methods. In the first ablation, we set $r$ to 0 so that there is no class regularization in DRL (``r = 0''). The prediction then follows the form in Eq.~\ref{eq:dratio}. In the second ablation, we set the density ratios to 1 instead of calculating it to mute the differentiable density ratio estimation in our method so that there is no representation level conservativeness (``R = 1''). When one method is existent, the performance is not degraded but also not boosted. However, DRST achieves the best results when both components are present, showing faster convergence and better performance. 

 \paragraph{Covariate shift:} Fig.~\ref{fig:drst}(c) shows how well the covariate shift assumption holds over the training process. We calculate $P_s(\bm\phi(\bm{x}))/P_t(\bm{\phi}(\bm{x})) - P_s(\bm{\phi}(\bm{x}), \bm{y})/P_t(\bm{\phi}(\bm{x}),\bm{y})$ using discriminative density ratio estimators (per class) as a proxy of covariate shift as it becomes 0 when covariate shift holds. We can see that the gap decreases with self-training, showing that even though covariate shift may not hold in the beginning, self-training helps to promote this assumption with better aligned domains and more discriminative feature distributions. Note that our model is tailored and effective to covariate shift no matter other shifts (e.g. label shift) exist or not. For example, we find label shift exists in VisDA2017 and our method still shows effectiveness: we calculate the division of the number of source samples over the number of target samples for each class, the numbers range from 1.42 to 4.23, indicating the existence of label shift in the dataset. Seeing the co-existence of multiple shifts and especially covariate shift, we believe our method is practical. This in fact has also well been verified by our real-world image experiments beyond covariate shift.

 \paragraph{Improvement on hard examples:} Fig.~\ref{fig:drst_d} demonstrates that compared to the baselines, DRST achieves larger performance gain on target samples with smaller density ratios. Recall that data is not well-represented in the source domain and is regarded as visually hard when the density ratio $P_{\text{s}}(\bm{x})/P_{\text{t}}(\bm{x})$ is small (Fig.~\ref{fig:network}(b)). Therefore, DRST provides more robust performance on harder examples.

 \paragraph{Density ratios:} Our density ratios are estimated from a differentiable domain classifier and is not guaranteed to match the true density ratios. However, our density ratios are interpretable as they reflect  the closeness of a sample to the two domains and benefit the downstream tasks. In Fig.~\ref{fig:network}(b), a harder example obtains a lower density ratio due to its vague shape (more examples are in the appendix).
 Moreover, the magnitude of our estimated density ratios are modest in general within the range of $[0.1, 10]$ (instead of approaching $0$ or infinity) due to the regularization by the target task classification network's learning signals.


 \paragraph{Improved attention:} We visualize the model attention of DRST using Grad-CAM~\cite{selvaraju2017grad} and compare with the CBST and CRST baselines in the appendix. The results show that DRST renders improved attention with better object coverage, presenting a better reception field.

\begin{table}[tb!]
    \centering
    \resizebox{0.48\textwidth}{!}{
    \begin{tabular}{ccccc}
    \toprule
    Dataset & CIFAR10 & STL10  & MNIST & \text{SVHN}\\
    \midrule 
    Fixmatch & 91.60 & 59.61  &99.43 & 26.50 \\
    \textbf{DRSSL} (Ours) & \textbf{95.17} & \textbf{69.38} & \textbf{99.46} & \textbf{30.96} \\
    \bottomrule
    \end{tabular}
    }
    \caption{Accuracy of cross-domain SSL in comparison with the peer method Fixmatch~\protect\cite{sohn2020fixmatch}.}
    \label{tab:ssl}
\end{table}

\subsection{Distributionally Robust Cross-Domain SSL}
To introduce domain gaps into semi-supervised learning, we choose the emerging cross-domain semi-supervised learning (CDSSL)~\cite{Yu2019UnsupervisedDA} setting. Specifically, CDSSL aims to use few labeled source training examples and many unlabeled target training examples to predict on source testing examples and target testing examples. Two pairs of source and target domains are used: Source: CIFAR10 / Target: STL10 and Source: MNIST / Target: SVHN. Under the single domain setting, FixMatch trains the model using source labeled data and source unlabeled data. However, under the cross-domain setting, we train the model using source labeled data and target unlabeled data. 

Table \ref{tab:ssl} shows that DRL-powered SSL method improves the Fixmatch baseline significantly on CDSSL tasks. Note that our setting is different from UDA where source labeled data is abundant. SSL focuses on using unlabeled augmented data to learn from few labeled data. In the CIFAR10 to STL10 case, we only have 4k labeled source data. For MNIST to SVHN, we have 40k labeled source data. The results show that DRL is beneficial for generating high-quality pseudo supervision for unlabeled data under the cross-domain SSL setting. More details of the experiments are provided in the appendix.

\section{Conclusion}
This paper studied uncertainty estimation under distribution shift with the distributionally robust learning framework. We show that density estimation can be integrated into the learning process by using a domain classifier. We propose differentiable density ratio estimation and develop end-to-end training techniques for our method. Using DRL's more calibrated model confidence helps to generate better pseudo-labels for self-training in UDA and cross-domain SSL. 

We also empirically show that the density ratios learned from our domain classifier reflect the hardness of an image, showing a positive correlation with the human selection frequencies. Future work involves relaxing the assumptions made and study different shifts in the DRL framework, which is in fact a limitation of our work.

\section*{Acknowledgments}
Haoxuan Wang started the work during the Summer Undergraduate Research Fellowship Program in Caltech. Anqi Liu was supported by the PIMCO and Amazon Postdoc Fellowship in Caltech and the Amazon-JHU AI2AI faculty award in JHU. Junchi Yan was supported by NSFC 62222607.

\bibliographystyle{named}
\bibliography{ijcai23}

\begin{thebibliography}{}

\bibitem[\protect\citeauthoryear{Antifakos \bgroup \em et al.\egroup
  }{2005}]{10.1145/1085777.1085780}
Stavros Antifakos, Nicky Kern, Bernt Schiele, and Adrian Schwaninger.
\newblock Towards improving trust in context-aware systems by displaying system
  confidence.
\newblock In {\em International Conference on Human Computer Interaction with
  Mobile Devices and Services}, 2005.

\bibitem[\protect\citeauthoryear{Bickel \bgroup \em et al.\egroup
  }{2007}]{bickel2007discriminative}
Steffen Bickel, Michael Br{\"u}ckner, and Tobias Scheffer.
\newblock Discriminative learning for differing training and test
  distributions.
\newblock In {\em ICML}, 2007.

\bibitem[\protect\citeauthoryear{Blundell \bgroup \em et al.\egroup
  }{2015}]{blundell2015weight}
Charles Blundell, Julien Cornebise, Koray Kavukcuoglu, and Daan Wierstra.
\newblock Weight uncertainty in neural network.
\newblock In {\em ICML}, 2015.

\bibitem[\protect\citeauthoryear{Brier}{1950}]{brier1950verification}
Glenn~W Brier.
\newblock Verification of forecasts expressed in terms of probability.
\newblock {\em Monthly weather review}, 1950.

\bibitem[\protect\citeauthoryear{Che \bgroup \em et al.\egroup }{2021}]{dvn}
Tong Che, Xiaofeng Liu, Site Li, Yubin Ge, Ruixiang Zhang, Caiming Xiong, and
  Yoshua Bengio.
\newblock Deep verifier networks: Verification of deep discriminative models
  with deep generative models.
\newblock {\em Proceedings of the AAAI Conference on Artificial Intelligence},
  35(8):7002--7010, May 2021.

\bibitem[\protect\citeauthoryear{Chen \bgroup \em et al.\egroup
  }{2020a}]{chen2020angular}
Beidi Chen, Weiyang Liu, Zhiding Yu, Jan Kautz, Anshumali Shrivastava, Animesh
  Garg, and Anima Anandkumar.
\newblock Angular visual hardness.
\newblock In {\em ICML}, 2020.

\bibitem[\protect\citeauthoryear{Chen \bgroup \em et al.\egroup
  }{2020b}]{chen2020automated}
Wuyang Chen, Zhiding Yu, Zhangyang Wang, and Animashree Anandkumar.
\newblock Automated synthetic-to-real generalization.
\newblock In {\em ICML}, 2020.

\bibitem[\protect\citeauthoryear{Coates \bgroup \em et al.\egroup
  }{2011}]{stl10}
Adam Coates, Andrew Ng, and Honglak Lee.
\newblock An analysis of single-layer networks in unsupervised feature
  learning.
\newblock In {\em AISTATS}, 2011.

\bibitem[\protect\citeauthoryear{Deng \bgroup \em et al.\egroup
  }{2009}]{2009ImageNet}
Jia Deng, Wei Dong, Richard Socher, Li-Jia Li, Kai Li, and Li~Fei-Fei.
\newblock Imagenet: A large-scale hierarchical image database.
\newblock In {\em CVPR}, 2009.

\bibitem[\protect\citeauthoryear{Fathony \bgroup \em et al.\egroup
  }{2016}]{fathony2016adversarial}
Rizal Fathony, Anqi Liu, Kaiser Asif, and Brian Ziebart.
\newblock Adversarial multiclass classification: A risk minimization
  perspective.
\newblock In {\em NIPS}, 2016.

\bibitem[\protect\citeauthoryear{Gal and Ghahramani}{2016}]{gal2016dropout}
Yarin Gal and Zoubin Ghahramani.
\newblock Dropout as a bayesian approximation: Representing model uncertainty
  in deep learning.
\newblock In {\em ICML}, 2016.

\bibitem[\protect\citeauthoryear{Ganin \bgroup \em et al.\egroup
  }{2016}]{ganin2016domain}
Yaroslav Ganin, Evgeniya Ustinova, Hana Ajakan, Pascal Germain, Hugo
  Larochelle, Fran{\c{c}}ois Laviolette, Mario Marchand, and Victor Lempitsky.
\newblock Domain-adversarial training of neural networks.
\newblock {\em The journal of machine learning research}, 2016.

\bibitem[\protect\citeauthoryear{Gr{\"u}nwald \bgroup \em et al.\egroup
  }{2004}]{grunwald2004game}
Peter~D Gr{\"u}nwald, A~Philip Dawid, et~al.
\newblock Game theory, maximum entropy, minimum discrepancy and robust bayesian
  decision theory.
\newblock {\em Annals of statistics}, 2004.

\bibitem[\protect\citeauthoryear{Guo \bgroup \em et al.\egroup
  }{2017}]{guo2017calibration}
Chuan Guo, Geoff Pleiss, Yu~Sun, and Kilian~Q Weinberger.
\newblock On calibration of modern neural networks.
\newblock In {\em ICML}, 2017.

\bibitem[\protect\citeauthoryear{Han \bgroup \em et al.\egroup
  }{2019}]{han2019unsupervised}
Ligong Han, Yang Zou, Ruijiang Gao, Lezi Wang, and Dimitris Metaxas.
\newblock Unsupervised domain adaptation via calibrating uncertainties.
\newblock In {\em CVPR Workshops}, volume~9, 2019.

\bibitem[\protect\citeauthoryear{He \bgroup \em et al.\egroup }{2016}]{resnet}
Kaiming He, Xiangyu Zhang, Shaoqing Ren, and Jian Sun.
\newblock Deep residual learning for image recognition.
\newblock In {\em CVPR}, 2016.

\bibitem[\protect\citeauthoryear{Hu \bgroup \em et al.\egroup
  }{2018}]{hu2018does}
Weihua Hu, Gang Niu, Issei Sato, and Masashi Sugiyama.
\newblock Does distributionally robust supervised learning give robust
  classifiers?
\newblock In {\em ICML}, 2018.

\bibitem[\protect\citeauthoryear{Khan \bgroup \em et al.\egroup
  }{2019}]{khan2019deep}
Haidar Khan, Lara Marcuse, and B{\"u}lent Yener.
\newblock Deep density ratio estimation for change point detection.
\newblock {\em arXiv preprint arXiv:1905.09876}, 2019.

\bibitem[\protect\citeauthoryear{Krizhevsky \bgroup \em et al.\egroup
  }{2012}]{alexnet}
Alex Krizhevsky, Ilya Sutskever, and Geoffrey~E Hinton.
\newblock Imagenet classification with deep convolutional neural networks.
\newblock {\em NIPS}, 2012.

\bibitem[\protect\citeauthoryear{Kumar \bgroup \em et al.\egroup
  }{2019}]{verifiedUC}
Ananya Kumar, Percy~S Liang, and Tengyu Ma.
\newblock Verified uncertainty calibration.
\newblock In H.~Wallach, H.~Larochelle, A.~Beygelzimer, F.~d\textquotesingle
  Alch\'{e}-Buc, E.~Fox, and R.~Garnett, editors, {\em Advances in Neural
  Information Processing Systems}, volume~32. Curran Associates, Inc., 2019.

\bibitem[\protect\citeauthoryear{Kumar \bgroup \em et al.\egroup
  }{2020}]{kumar2020understanding}
Ananya Kumar, Tengyu Ma, and Percy Liang.
\newblock Understanding self-training for gradual domain adaptation.
\newblock In {\em ICML}, 2020.

\bibitem[\protect\citeauthoryear{Lecun and Bottou}{1998}]{mnist}
Y.~Lecun and L.~Bottou.
\newblock Gradient-based learning applied to document recognition.
\newblock {\em Proceedings of the IEEE}, 1998.

\bibitem[\protect\citeauthoryear{Lee and Lee}{2020}]{lee2020model}
JoonHo Lee and Gyemin Lee.
\newblock Model uncertainty for unsupervised domain adaptation.
\newblock In {\em 2020 IEEE International Conference on Image Processing
  (ICIP)}. IEEE, 2020.

\bibitem[\protect\citeauthoryear{Li and Hoiem}{2020}]{li2020improving}
Zhizhong Li and Derek Hoiem.
\newblock Improving confidence estimates for unfamiliar examples.
\newblock In {\em CVPR}, 2020.

\bibitem[\protect\citeauthoryear{Liu and Ziebart}{2014}]{liu2014robust}
Anqi Liu and Brian Ziebart.
\newblock Robust classification under sample selection bias.
\newblock In {\em NIPS}, 2014.

\bibitem[\protect\citeauthoryear{Liu and Ziebart}{2017}]{liu2017robust}
Anqi Liu and Brian~D Ziebart.
\newblock Robust covariate shift prediction with general losses and feature
  views.
\newblock {\em arXiv preprint arXiv:1712.10043}, 2017.

\bibitem[\protect\citeauthoryear{Liu \bgroup \em et al.\egroup
  }{2020}]{liu2020robust}
Anqi Liu, Guanya Shi, Soon-Jo Chung, Anima Anandkumar, and Yisong Yue.
\newblock Robust regression for safe exploration in control.
\newblock In {\em Learning for Dynamics and Control}, 2020.

\bibitem[\protect\citeauthoryear{Long \bgroup \em et al.\egroup
  }{2018}]{long2018conditional}
Mingsheng Long, Zhangjie Cao, Jianmin Wang, and Michael~I Jordan.
\newblock Conditional adversarial domain adaptation.
\newblock In {\em NeurIPS}, 2018.

\bibitem[\protect\citeauthoryear{Najafi \bgroup \em et al.\egroup
  }{2019}]{najafi2019robustness}
Amir Najafi, Shin-ichi Maeda, Masanori Koyama, and Takeru Miyato.
\newblock Robustness to adversarial perturbations in learning from incomplete
  data.
\newblock In {\em NeurIPS}, 2019.

\bibitem[\protect\citeauthoryear{Nakka \bgroup \em et al.\egroup
  }{2020}]{nakka2020chance}
Yashwanth~Kumar Nakka, Anqi Liu, Guanya Shi, Anima Anandkumar, Yisong Yue, and
  Soon-Jo Chung.
\newblock Chance-constrained trajectory optimization for safe exploration and
  learning of nonlinear systems.
\newblock {\em IEEE Robotics and Automation Letters}, 2020.

\bibitem[\protect\citeauthoryear{Netzer \bgroup \em et al.\egroup
  }{2011}]{svhn}
Yuval Netzer, Tao Wang, Adam Coates, Alessandro Bissacco, Bo~Wu, and Andrew~Y.
  Ng.
\newblock Reading digits in natural images with unsupervised feature learning.
\newblock In {\em NIPS Workshop on Deep Learning and Unsupervised Feature
  Learning 2011}, 2011.

\bibitem[\protect\citeauthoryear{Nixon \bgroup \em et al.\egroup
  }{2019}]{nixon2019measuring}
Jeremy Nixon, Michael~W Dusenberry, Linchuan Zhang, Ghassen Jerfel, and Dustin
  Tran.
\newblock Measuring calibration in deep learning.
\newblock In {\em CVPR Workshops}, 2019.

\bibitem[\protect\citeauthoryear{Park \bgroup \em et al.\egroup
  }{2020}]{park2020calibrated}
Sangdon Park, Osbert Bastani, James Weimer, and Insup Lee.
\newblock Calibrated prediction with covariate shift via unsupervised domain
  adaptation.
\newblock In {\em AISTATS}, 2020.

\bibitem[\protect\citeauthoryear{Peng \bgroup \em et al.\egroup
  }{2017}]{visda2017}
Xingchao Peng, Ben Usman, Neela Kaushik, Judy Hoffman, Dequan Wang, and Kate
  Saenko.
\newblock Visda: The visual domain adaptation challenge, 2017.

\bibitem[\protect\citeauthoryear{Pereyra \bgroup \em et al.\egroup
  }{2017}]{pereyra2017regularizing}
Gabriel Pereyra, George Tucker, Jan Chorowski, {\L}ukasz Kaiser, and Geoffrey
  Hinton.
\newblock Regularizing neural networks by penalizing confident output
  distributions.
\newblock {\em arXiv preprint arXiv:1701.06548}, 2017.

\bibitem[\protect\citeauthoryear{Platt and
  others}{1999}]{platt1999probabilistic}
John Platt et~al.
\newblock Probabilistic outputs for support vector machines and comparisons to
  regularized likelihood methods.
\newblock {\em Advances in large margin classifiers}, 1999.

\bibitem[\protect\citeauthoryear{Qiu \bgroup \em et al.\egroup
  }{2021}]{Qiu2021SourcefreeDA}
Zhen Qiu, Yifan Zhang, Hongbin Lin, Shuaicheng Niu, Yanxia Liu, Qing Du, and
  Mingkui Tan.
\newblock Source-free domain adaptation via avatar prototype generation and
  adaptation.
\newblock {\em ArXiv}, abs/2106.15326, 2021.

\bibitem[\protect\citeauthoryear{Recht \bgroup \em et al.\egroup
  }{2019}]{2019imagenetv2}
Benjamin Recht, Rebecca Roelofs, Ludwig Schmidt, and Vaishaal Shankar.
\newblock Do {I}mage{N}et classifiers generalize to {I}mage{N}et?
\newblock In {\em Proceedings of the 36th International Conference on Machine
  Learning}, volume~97 of {\em Proceedings of Machine Learning Research}, pages
  5389--5400. PMLR, 09--15 Jun 2019.

\bibitem[\protect\citeauthoryear{Saenko \bgroup \em et al.\egroup
  }{2010}]{office31}
Kate Saenko, Brian Kulis, Mario Fritz, and Trevor Darrell.
\newblock Adapting visual category models to new domains.
\newblock In {\em ECCV}. Springer, 2010.

\bibitem[\protect\citeauthoryear{Saito \bgroup \em et al.\egroup
  }{2018a}]{saito2018adversarial}
Kuniaki Saito, Yoshitaka Ushiku, Tatsuya Harada, and Kate Saenko.
\newblock Adversarial dropout regularization.
\newblock In {\em ICLR}, 2018.

\bibitem[\protect\citeauthoryear{Saito \bgroup \em et al.\egroup
  }{2018b}]{saito2018maximum}
Kuniaki Saito, Kohei Watanabe, Yoshitaka Ushiku, and Tatsuya Harada.
\newblock Maximum classifier discrepancy for unsupervised domain adaptation.
\newblock In {\em CVPR}, 2018.

\bibitem[\protect\citeauthoryear{Selvaraju \bgroup \em et al.\egroup
  }{2017}]{selvaraju2017grad}
Ramprasaath~R Selvaraju, Michael Cogswell, Abhishek Das, Ramakrishna Vedantam,
  Devi Parikh, and Dhruv Batra.
\newblock {Grad-CAM}: Visual explanations from deep networks via gradient-based
  localization.
\newblock In {\em ICCV}, 2017.

\bibitem[\protect\citeauthoryear{Shu \bgroup \em et al.\egroup
  }{2018}]{shu2018dirt}
Rui Shu, Hung~H Bui, Hirokazu Narui, and Stefano Ermon.
\newblock A dirt-t approach to unsupervised domain adaptation.
\newblock {\em ICLR}, 2018.

\bibitem[\protect\citeauthoryear{Simonyan and Zisserman}{2014}]{vgg19}
Karen Simonyan and Andrew Zisserman.
\newblock Very deep convolutional networks for large-scale image recognition.
\newblock {\em arXiv preprint arXiv:1409.1556}, 2014.

\bibitem[\protect\citeauthoryear{Snoek \bgroup \em et al.\egroup
  }{2019}]{snoek2019can}
Jasper Snoek, Yaniv Ovadia, Emily Fertig, Balaji Lakshminarayanan, Sebastian
  Nowozin, D~Sculley, Joshua Dillon, Jie Ren, and Zachary Nado.
\newblock Can you trust your model's uncertainty? evaluating predictive
  uncertainty under dataset shift.
\newblock In {\em NeurIPS}, 2019.

\bibitem[\protect\citeauthoryear{Sohn \bgroup \em et al.\egroup
  }{2020}]{sohn2020fixmatch}
Kihyuk Sohn, David Berthelot, Chun-Liang Li, Zizhao Zhang, Nicholas Carlini,
  Ekin~D. Cubuk, Alex Kurakin, Han Zhang, and Colin Raffel.
\newblock Fixmatch: Simplifying semi-supervised learning with consistency and
  confidence.
\newblock In {\em NeurIPS}, 2020.

\bibitem[\protect\citeauthoryear{Sugiyama \bgroup \em et al.\egroup
  }{2012}]{sugiyama2012density}
Masashi Sugiyama, Taiji Suzuki, and Takafumi Kanamori.
\newblock {\em Density ratio estimation in machine learning}.
\newblock Cambridge University Press, 2012.

\bibitem[\protect\citeauthoryear{Szegedy \bgroup \em et al.\egroup
  }{2016}]{szegedy2016rethinking}
Christian Szegedy, Vincent Vanhoucke, Sergey Ioffe, Jon Shlens, and Zbigniew
  Wojna.
\newblock Rethinking the inception architecture for computer vision.
\newblock In {\em CVPR}, 2016.

\bibitem[\protect\citeauthoryear{Tomsett \bgroup \em et al.\egroup
  }{2020}]{tomsett2020rapid}
Richard Tomsett, Alun Preece, Dave Braines, Federico Cerutti, Supriyo
  Chakraborty, Mani Srivastava, Gavin Pearson, and Lance Kaplan.
\newblock Rapid trust calibration through interpretable and uncertainty-aware
  ai.
\newblock {\em Patterns}, 2020.

\bibitem[\protect\citeauthoryear{Tzeng \bgroup \em et al.\egroup }{2017}]{adda}
Eric Tzeng, Judy Hoffman, Kate Saenko, and Trevor Darrell.
\newblock Adversarial discriminative domain adaptation.
\newblock In {\em CVPR}, 2017.

\bibitem[\protect\citeauthoryear{Venkateswara \bgroup \em et al.\egroup
  }{2017}]{officehome}
Hemanth Venkateswara, Jose Eusebio, Shayok Chakraborty, and Sethuraman
  Panchanathan.
\newblock Deep hashing network for unsupervised domain adaptation.
\newblock In {\em CVPR}, 2017.

\bibitem[\protect\citeauthoryear{Wang \bgroup \em et al.\egroup
  }{2020}]{wang2020transferable}
Ximei Wang, Mingsheng Long, Jianmin Wang, and Michael~I Jordan.
\newblock Transferable calibration with lower bias and variance in domain
  adaptation.
\newblock {\em arXiv preprint arXiv:2007.08259}, 2020.

\bibitem[\protect\citeauthoryear{Xu \bgroup \em et al.\egroup
  }{2021}]{Xu2021CDTransCT}
Tongkun Xu, Weihua Chen, Pichao Wang, Fan Wang, Hao Li, and Rong Jin.
\newblock Cdtrans: Cross-domain transformer for unsupervised domain adaptation.
\newblock {\em ArXiv}, abs/2109.06165, 2021.

\bibitem[\protect\citeauthoryear{Yu \bgroup \em et al.\egroup
  }{2019}]{Yu2019UnsupervisedDA}
Fuxun Yu, Di~Wang, Yinpeng Chen, Nikolaos Karianakis, Pei Yu, Dimitrios
  Lymberopoulos, and Xiang Chen.
\newblock Unsupervised domain adaptation for object detection via cross-domain
  semi-supervised learning.
\newblock {\em ArXiv}, abs/1911.07158, 2019.

\bibitem[\protect\citeauthoryear{Zou \bgroup \em et al.\egroup
  }{2018}]{zou2018domain}
Yang Zou, Zhiding Yu, BVK Kumar, and Jinsong Wang.
\newblock Unsupervised domain adaptation for semantic segmentation via
  class-balanced self-training.
\newblock In {\em ECCV}, 2018.

\bibitem[\protect\citeauthoryear{Zou \bgroup \em et al.\egroup
  }{2019}]{zou2019confidence}
Yang Zou, Zhiding Yu, Xiaofeng Liu, BVK Kumar, and Jinsong Wang.
\newblock Confidence regularized self-training.
\newblock In {\em ICCV}, 2019.

\end{thebibliography}


\end{document}